\documentclass[10pt,twocolumn,letterpaper]{article}

\usepackage{iccv}
\usepackage{times}
\usepackage{epsfig}
\usepackage{graphicx}
\usepackage{amsmath,amssymb} 
\usepackage{color}
\usepackage{multirow}
\usepackage{booktabs}
\usepackage{subfigure}

\usepackage{float}

\iccvfinalcopy 

\def\iccvPaperID{****} 

\author{\parbox{16cm}{\centering
    {\large Siyang Song$^1$, Enrique S\'anchez-Lozano$^1$,  Linlin Shen$^2$, Alan Johnston$^3$ and  Michel Valstar$^1$}\\
    {\normalsize
    $^1$ School of Computer Science, University of Nottingham, UK\\
    $^2$ College of Computer Science and Software Engineering, Shenzhen University, China \\
    $^3$ School of Psychology, University of Nottingham, UK}
    }
}

\begin{document}

\title{Inferring Dynamic Representations of Facial Actions from a Still Image}

\maketitle

\begin{abstract}
\noindent Facial actions are spatio-temporal signals by nature, and therefore their modeling is crucially dependent on the availability of temporal information. In this paper, we focus on inferring such temporal dynamics of facial actions when no explicit temporal information is available, i.e. from still images. We present a novel approach to capture multiple scales of such temporal dynamics, with an application to facial Action Unit (AU) intensity estimation and dimensional affect estimation. In particular, 1) we propose a framework that infers a dynamic representation (DR) from a still image, which captures the bi-directional flow of time within a short time-window centered at the input image; 2) we show that we can train our method without the need of explicitly generating target representations, allowing the network to represent dynamics more broadly; and 3) we propose to apply a multiple temporal scale approach that infers DRs for different window lengths (MDR) from a still image. We empirically validate the value of our approach on the task of frame ranking, and show how our proposed MDR attains state of the art results on BP4D for AU intensity estimation and on SEMAINE for dimensional affect estimation, using only still images at test time. 
\end{abstract}


\section{Introduction}

\begin{figure}
\centering
\includegraphics[width=9.2cm]{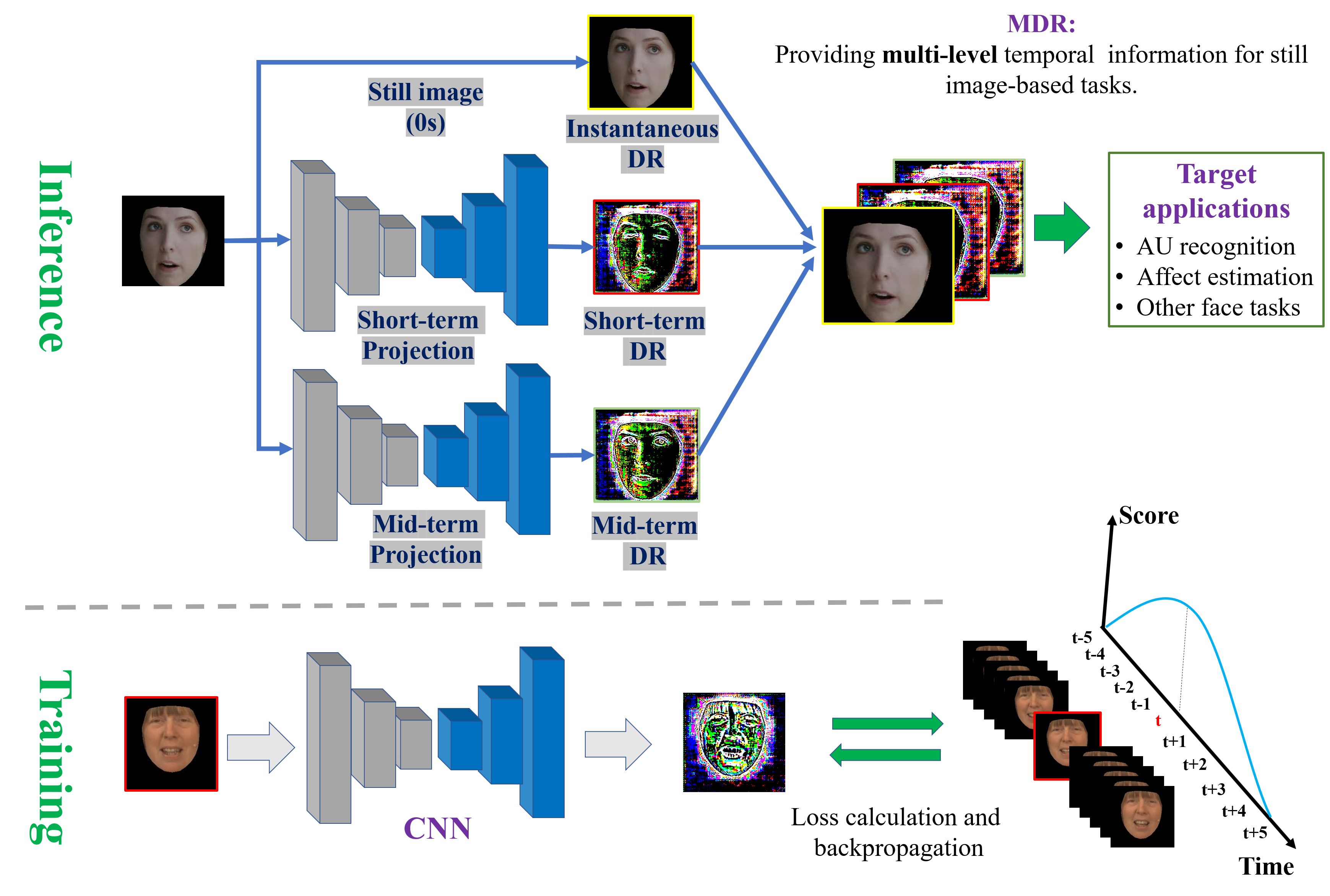}
\caption{We propose a novel approach to the modeling of temporal face dynamics from still images (top). Our approach can be used to infer several time-length dynamics, and further combined to enhance face-related tasks such as AU intensity estimation and dimensional affect estimation. During training (bottom), a set of videos is used to learn the DRs, without explicitly generating a fixed set of target representations.}
\label{fig:first}
\end{figure}


\noindent Temporal dynamics are an important source of information for video-based face analysis. In recent years, many methods have been proposed to exploit that for tasks where the temporal information correlates over time with the target signals~\cite{chu2017learning,jaiswal2016deep,zhang2017facial,xu2017microexpression}. The temporal modeling can be accomplished either by generating a single set of features from multiple consecutive frames at a time (early modeling of temporal dynamics~\cite{zhang2017facial}) or by using memory-based models, such as Recurrent Neural Networks (RNNs) or Markov Models (late modeling of temporal dynamics~\cite{chu2017learning}), or by a combination of both~\cite{jaiswal2016deep}. Recently, there is an increasing interest in the early modeling of temporal dynamics, as these can be straightforwardly used in simple CNN-based architectures. Among these methods, the dynamic image~\cite{bilen2017action} has achieved promising results at the task of summarizing short-term motion in a fixed-size representation, for the task of action recognition. 

While the use of temporal dynamics is desirable for facial behaviour-related tasks, it is often not available. For example, in many applications across domains ranging from medicine to marketing, the analysis of facial expressions or emotions is required from  still images. Without temporal information, the performance of state of the art methods for facial expression recognition or affect estimation degrade substantially~\cite{Kollias2019,jaiswal2016deep}. While some works have been proposed to anticipate or predict the next frame from a still image or from a video~\cite{Pintea2014DejaVM, walker2016uncertain}, none of existing works have attempted to model the dynamics of facial expressions from a still image. 

In this paper, \textit{we want to generate a dynamic representation that summarizes motion, but that can be inferred from still images}. To the best of our knowledge, this is the first work that brings the advantages of DRs at summarizing sequences to scenarios where only still images are given. Even though some early works exist that summarize temporal dynamics from image sequences~\cite{bilen2017action}, or attempt to predict or anticipate motion from still images~\cite{Pintea2014DejaVM, walker2016uncertain}, to the best of our knowledge, there are no works trying to infer a DR from a still image. In this paper, we are interested in \textit{inferring} the temporal dynamics from a previously unseen face image, and in the use of this information to further enhance the performance of Action Unit (AU) intensity estimation and dimensional affect estimation networks. To this end, we propose an image-to-image translation approach where a network is trained to generate a Dynamic Representation (DR) from a given image. Inspired by the dynamic image~\cite{bilen2017action}, our representation is formulated as a kernel that, when projected onto the adjacent frames, can sort them in time. It is therefore designed to be a single three-channel spatial data structure (similar to an RGB image), tasked with summarizing the motion that surrounds a given frame. This representation can be directly used in CNN-based networks for the tasks of AU intensity estimation and dimensional affect estimation.

\begin{figure*}
\centering
\includegraphics[width=16.5cm]{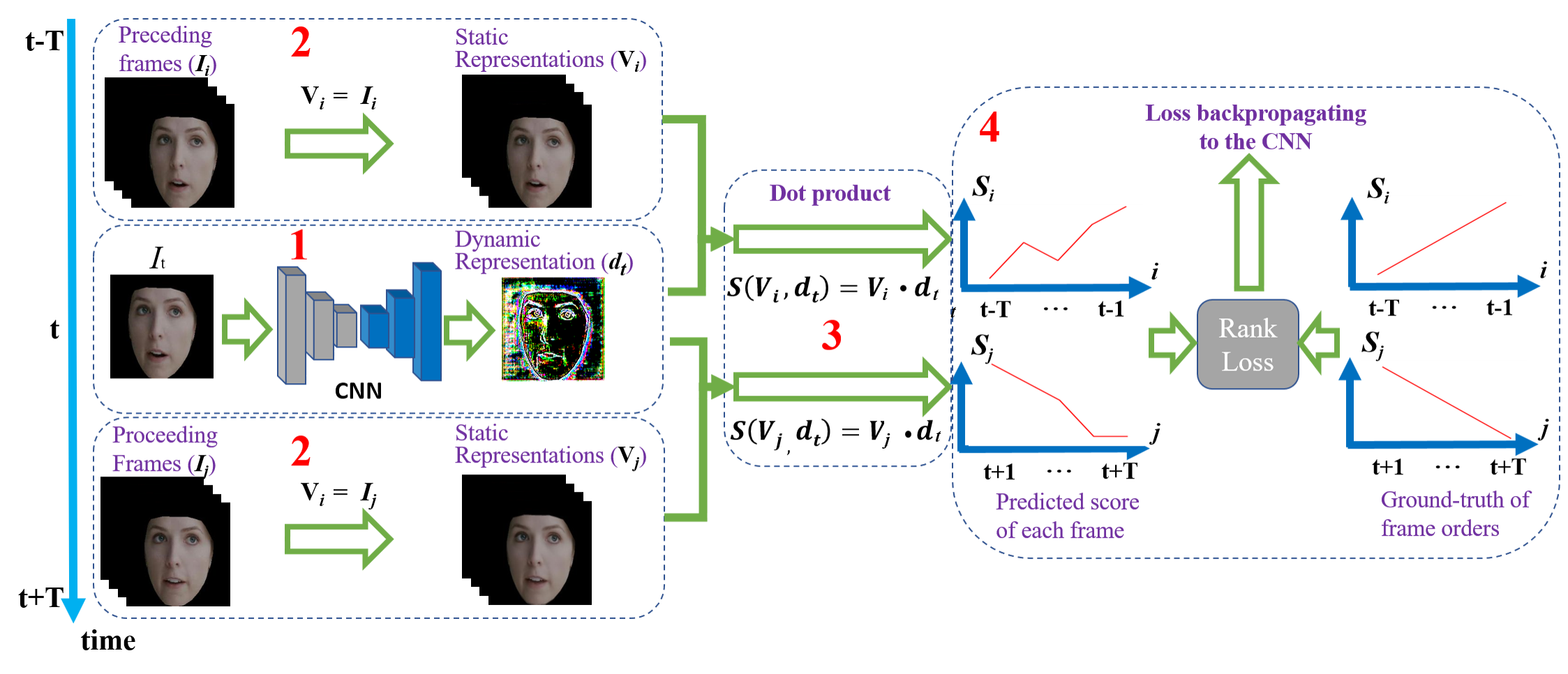}
\caption{ During training, we are given a set of sequences from which we can have access to the adjacent frames of a given image $I_t$. In (1), the given image $I_t$ is forwarded to the network we aim to learn, that produces a DR $d_t$. We can measure the ranking capabilities of $d_t$ by projecting it onto the preceding and proceeding frames (2). To rank the frames, we compute the difference between pair-wise scores, each computed as a dot product between the generated DR and the corresponding preceding or proceeding frame (3). These scores are used to compute a Rank Loss, which allow us to measure the extent of which the current $d_t$ is correctly ranking the frames within the sequence. We can backpropagate the Rank Loss w.r.t. the parameters of the network that has produced the DR $d_t$ (4). This way, the network not only learns to produce a correct representation $d_t$, but also contributes to define it.}
\label{fig:diagram}
\end{figure*}

During training, we are given a set of preceding and proceeding frames for each face image (i.e. sequences), from which the temporal evolution of adjacent frames can be learned in a self-supervised manner without using target representations (see Fig.~\ref{fig:diagram}). The network is then trained to generate a representation that, when projected onto each adjacent frame within a given window, is capable of sorting them in time. In other words, \textit{we do not compute a set of target representations to learn our network}. In addition, we note that the temporal symmetry of facial actions could yield ambiguities when sorting frames in a strictly ascending order, and propose to sort frames relative to their distance to the central frame. We first validate empirically that the learned representation does have the capacity to sort adjacent frames in the temporal domain, thus illustrating its ability to capture short-term dependencies. Then, we show that our proposed approach can generate representations that are highly suitable for both AU intensity estimation and dimensional affect estimation. In particular, we propose a multi-level dynamic representation approach (see Fig.~\ref{fig:first}), that combines DRs generated for different temporal scales. We show this approach suffices to attain state of the art results in both tasks. 
Our contributions can be summarized as follows:
\begin{itemize}
    \item We propose an image-to-image translation network, tasked with \textit{inferring} a dynamic representation of a given still image, designed to summarize the short-term motion surrounding it.
    
    \item We propose to train the network with a Rank Loss, enforcing the generated representations to rank both past and future frames according to their relative distance to the central frame. This way, the network not only learns to map an image to a corresponding representation, but also contributes to define it. 
    
    \item We show how the inferred representations effectively summarize motion, and show how their use in combination with a given frame reaches state of the art results in the tasks of facial Action Unit (AU) intensity estimation and dimensional affect estimation. 
\end{itemize}

\section{Related Work}
\noindent This Section reviews the closely related work, which we define as works related to the temporal modeling of facial motion and  motion estimation from still images (our main goal), as well as image-to-image translation, image-based dynamic representations and self-supervised learning. 

\textbf{Temporal modeling of facial expressions} Exploiting the temporal modeling of facial expressions on video sequences is a longstanding problem in Computer Vision. Some works have proposed to summarize short-term motion at the feature level, extending hand-crafted features to what is known as Three Orthogonal Planes (TOP)~\cite{chen2018facial,jiang2014dynamic}. Other works have exploited the use of a Fourier Transform~\cite{song2018human}, or spatio-temporal convolution~\cite{jaiswal2016deep, zhang2017facial}. The majority of related work focus on using recurrent or latent-based models, in particular Recurrent Neural Networks (RNNs~\cite{chu2017learning,jaiswal2016deep,zhang2017facial,mohammad2017facial,fan2016video,Kollias2019}).


\textbf{Motion Estimation} Our work is related to motion prediction, where the goal is to infer motion from either still images or sequences. In this sense, the goal is to predict \textit{what is going to happen next}. Some works have tackled this problem by predicting optical flow from still images~\cite{Pintea2014DejaVM,walker2015dense}. Others have proposed to infer the next frame to follow a preceding video sequence~\cite{chen2017video,xue2018visual}. In particular, \cite{rodriguez2018action} proposed to infer the next dynamic image, as it better correlates with the preceding frames. These methods do not attempt to summarize motion, but rather predict the most likely frame to follow a given image or image sequence. 


\textbf{Image-to-image translation} Our work can be viewed as  image-to-image translation, where a dynamic representation (a 3-channel image in our case) is generated from an input image. Works in image-to-image translation generally attempt to modify an input image to generate an output according to a target attribute or style, and thus do not have as a goal to \textit{infer} any information \emph{from} the input image~\cite{choi2018stargan,Isola2017ImagetoImageTW,long2015fully,wang2018high,yi2017dualgan,zhang2016colorful, pumarola2018ganimation}. These approaches generally rely on the use of Generative Adversarial Networks (GANs)~\cite{goodfellow2014generative}, or any of its extensions~\cite{choi2018stargan,Isola2017ImagetoImageTW,yi2017dualgan}. GANs are a powerful tool to capture the target distribution, enforcing the networks to produce plausible outputs. However, as we shall see, we will not be using explicit target representations to learn our network, and therefore the use of GANs is not a suitable tool to learn the dynamic representations. 

\textbf{Dynamic Representation} The basis of our work is referred to as dynamic representation. A dynamic representation is built so that it can rank order frames according to their position in a temporal sequence. The reasoning behind such an abstract representation is that if a representation has the power to rank all frames according to their temporal position in an image sequence, then it is a good descriptor of it, and thus can be used for machine learning tasks that require this temporal information. This hypothesis was validated for the task of human action recognition~\cite{bilen2017action,fernando2017rank}. The dynamic representation (referred to as \textit{dynamic image} in \cite{bilen2017action}) was first presented as a short-term feature descriptor of image sequences. To obtain a dynamic image, one needs to \textit{learn} it at test time from the set of frames that make up a sequence using e.g. RankSVM~\cite{smola2004tutorial}. The use of RankSVM was further extended and converted into a pooling layer~\cite{fernando2017rank}. As we shall see, our network will be able to generate a dynamic representation \textit{from still images}, that effectively summarizes not only past frames, but also future frames. Finally, it is worth mentioning that before the dynamic image, other methods were proposed to learn dynamic representations from image sequences, such as optical flow or Motion History Image~\cite{bobick2001recognition}.

\textbf{Self-supervised Learning} In this paper we propose to learn without explicitly generating target representations. Instead, we will make use of a proxy loss function, called a Rank Loss, to train our network in a self-supervised manner. Self-supervised learning avoids the need of explicit target data, and instead explores the structure of the training data to supervise the training process, using e.g. temporal relations or semantic structures~\cite{doersch2015unsupervised}. Some works on self-supervised learning have already used the temporal order of video frames to train networks, aiming to learn video representations of asymmetric human actions ~\cite{fernando2017self,misra2016shuffle} or analyze temporal coherence ~\cite{jayaraman2016slow,goroshin2015unsupervised}. To the best of our knowledge, we are the first to propose the use of a Rank Loss function to learn a dynamic representation of facial expressions in a self-supervised manner.

\section{Proposed approach}
\noindent Our goal is to train a network that generates a simple dynamic representation (DR) from a single face image, summarizing the motion around it. This is possible because facial actions are constrained by anatomy and behaviour causes strong correlations between adjacent frames. In this Section, we first define the goal of the DRs. To do so, we introduce the challenges that facial expressions pose for the task of learning a representation that summarizes motion, and propose an alternative representation that overcomes them. Then, we show how this representation can be learned. We note that generating a set of target representations to be the basis of a one-to-one mapping, common in image-to-image translation methods, is suboptimal, and propose an alternative approach to allow the learning process to be self-supervised by the temporal correlation between frames in the training set. Finally, we show how to apply the trained models for the task of AU intensity estimation and dimensional affect estimation. The full learning process is shown in Fig.~\ref{fig:diagram}.



\begin{figure}[t!]
	\centering
		\includegraphics[width=8cm]{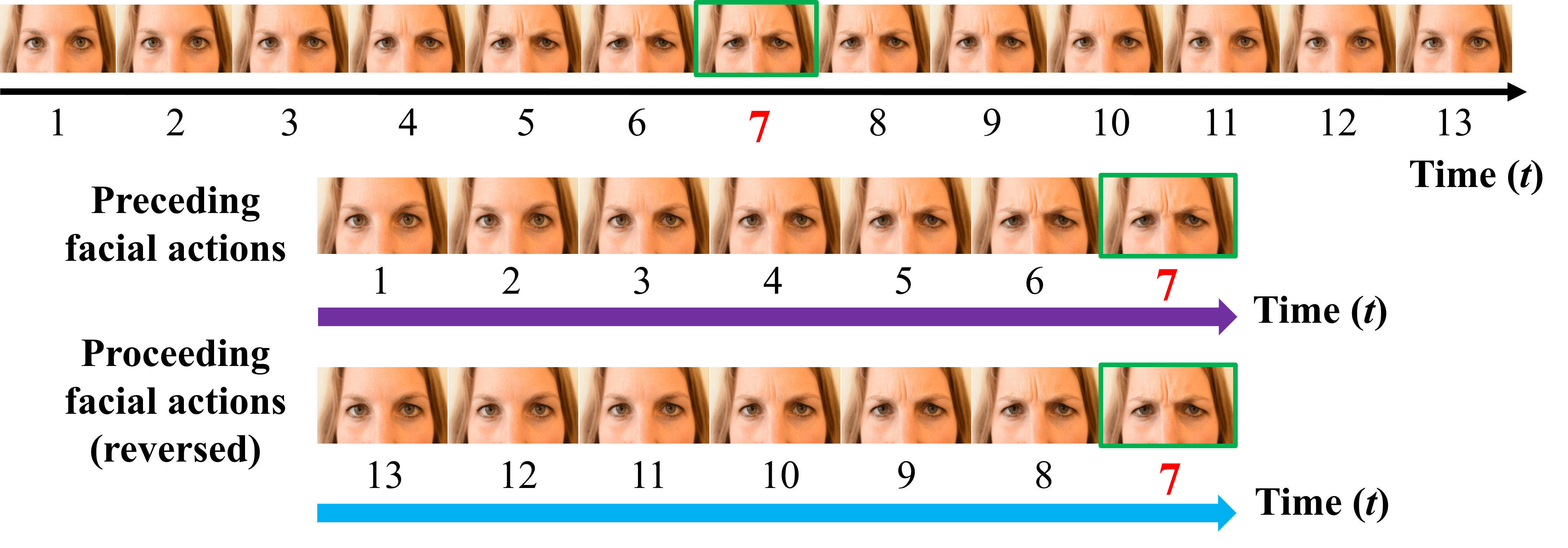}
	\caption{Temporal symmetry of facial motions. In the top row, a sequence of frames displaying a facial expression. With regards to the central frame $t=7$, we can see that the preceding frames would look alike the proceeding frames, when the latter are reverted on time, i.e. the temporal pattern evolved from $t=1$ to $t=7$ with the natural arrow of time would be similar to the temporal pattern evolved from $t=13$ to $t=7$, under a reverted arrow of time. 
    } \label{fig:illustrate}
\end{figure}


\subsection{Dynamic Representations}


\noindent As mentioned above, our goal is to train a network that can infer a DR from a still face image. The first step consists of defining what output the network is expected to return for a given image. To do so, we draw our attention to the dynamic image algorithm~\cite{bilen2017action}. The dynamic algorithm was presented as a kernel targeted with summarizing sequences of frames, with the goal of inferring the action on them. The main motivation behind generating a kernel that can rank frames relies on the assumption that such a representation should be a good descriptor of the sequence, as it encodes the temporal evolution of it. This assumption was empirically validated in~\cite{bilen2017action}. In our framework, we are instead interested in having a per-frame representation, and therefore the original dynamic algorithm seems not to be an adequate representation. We empirically validate in Sec.~\ref{ssec:ranking} that, when the dynamic image is used as a basis to learn the representation, along with a reconstruction loss, the performance degrades substantially. In addition, it is important to note that facial expressions display a \textbf{symmetric} temporal pattern. As shown in Fig.~\ref{fig:illustrate}, facial expressions are in many cases indistinguishable from their (temporally) reverted counterparts. In such cases, using the dynamic algorithm would incur in ambiguous representations for a given frame, according to whether it belongs to the ``activation" part of an expression, or whether it belongs to its ``deactivation" part. In order to overcome such limitation, we propose a DR that is targeted with ranking not only the preceding frames, but also the proceeding frames. In other words, the DR is chosen to be a kernel that can rank both past and future frames, based on their temporal positions relative to the given face image. This way, the modeling of symmetric patterns is addressed in a more efficient way.

Let $I_t \in \mathbb{R}^{m \times n}$ be the given face image, and let $I_{t-T}, I_{t-T+1}, \cdots, I_{t-1}$ and $I_{t+1}, I_{t+2}, \cdots, I_{t+T}$ be the frames corresponding to a window of $2T+1$ frames, centered at $I_t$. Let $V_a , a \in [t-T, t+T]$ be the static representation for the frame $a$. While in~\cite{bilen2017action} $V_a$ was defined as the cumulative feature representation of the frame $a$ (i.e. $V_a = \sum_{a'<a} \phi(I_{a'})$, with $\phi$ a feature representation), in this paper we directly choose $V_a$ to be the image itself, i.e. $V_a = I_a$, to avoid the feature representation to depend on other frames. Our goal is to generate a DR for frame $I_t$, namely $d_t$, with size equal to the static representation $V_a$, that can rank preceding and proceeding frames based on their relative temporal distance to $I_t$. The ranking of frames is performed by assigning a score to each, which is defined as the (Frobenius) inner product between the DR $d_t$ and the representation $V_a$, with $a \in [t-T, t+T]$. Mathematically speaking, the score for frame $a \in [t-T, t+T]$, assigned by the DR $d_t$ is defined as $S(d_t, V_a) = \langle d_t, V_a \rangle$. The scores are then used in an ordinal manner to sort the frames within the given window. In particular, we are interested in assigning ascending scores for frames $a < t$, and descending scores for $a > t$. This way, we can define the difference between the scores computed at time $a$ and $b$ as 

\begin{equation}
\label{eq:score}
\delta_{ab}(t) \doteq S(d_t, V_a ) - S(d_t, V_b)
\end{equation}
where $a$ is chosen to be closer to $t$ than $b$, with both $a$ and $b$ being either preceding or proceeding frames w.r.t. $t$. Then, our goal is to learn a DR $d_t$ that makes $\delta_{ab}(t) > 0$ for pairs $a,b$ corresponding to $a,b > t$ or $a,b < t$, with $| a - t | < | b - t |$. As we shall see in Sec.~\ref{ssec:learning}, we will only consider the actual value of $\delta_{ab}(t)$ when the pair $a,b$ has been incorrectly ranked, i.e. when $\delta_{ab}(t) < 0$. Thus, we can define our target representation as a $d_t$ that meets the following criteria:
\begin{equation}
\begin{aligned}
& \delta_{ab}(t) > 0 \quad \text{for} \quad \begin{cases}
    | a - t | < | b - t |       \\
    (a - t)(b - t) > 0  
  \end{cases} 
\end{aligned}
\label{eq:definition}
\end{equation}
It is important to remark that we compute the scores for the cases in which both frames are either before the current frame $t$, or after it. We are interested in computing ascending scores when $a,b < t$, and descending scores when $a,b > t$, and thus the cases where e.g. $a > t$ and $b < t$ would raise complex definitions.


\subsection{Learning DR with Rank Loss}
\label{ssec:learning}
\noindent We now focus on how to learn the DR described above. Recall that our ultimate goal is to learn a static-to-dynamic projection $f$ from a static face image $I_t$ to its DR $d_t$, i.e. $d_t = f(I_t)$. This representation is tasked with meeting the aforementioned criteria for all the training set of available frames. A priory, this could be accomplished by first generating the target representations from a sequence and then by training a network from the corresponding pairs, i.e. the centre image of the sequence and the target representations. In such case, one could use a reconstruction loss between the generated output and the corresponding target, so that the network learns to reproduce such a representation. However, we observe that when using a pre-defined representation to train the network, the generated outputs lack of generalization. We validate this empirically in Sec.~\ref{ssec:ranking}. In other words, the network is forced to minimize a reconstruction loss w.r.t. a fixed representation, and therefore it does not take into account the capabilities of the generated output at the task of ranking adjacent frames. That is to say, \textit{subtle errors in the reconstruction loss do not necessarily correlate with errors in the ranking of frames}. Instead, we want the network to also help design the DR. 

In particular, instead of generating target representations, we propose to learn the DR by enforcing the network to produce outputs that directly meet Eqn.~\ref{eq:definition}. More specifically, when the network generates an output for a given image, we project it onto the preceding and proceeding frames within a window of $N = 2T+1$ frames, and compute the pair-wise scores using Eqn.~\ref{eq:score} and Eqn.~\ref{eq:definition}. Then, in a similar fashion to that of the RankSVM algorithm, we only account for the error committed by the pairs that have been incorrectly ranked. In addition, we add a rank success factor $\theta$, to avoid small errors to be considered in the total loss. Mathematically speaking, let $I_t$ be the given frame, corresponding to the central image of a window of $N = 2T + 1$ frames. Let $d_t = f(I_t)$ be the output of the network for the given frame. We want the generated DR to minimize the following rank loss function:
\begin{equation}
\begin{split}
L_{f}(d_t) & = \gamma \times \|\mathrm{d_t}\|^2  - \varepsilon  \\
& +  \sum_{b=t-T}^{t-1} \sum_{a=b+1}^{t} \max (0, \theta - \delta_{ab}(t) ) \\
& +  \sum_{a=t}^{t+T-1} \sum_{b=a+1}^{T} \max (0, \theta - \delta_{ab}(t) )
\end{split}
\label{eq:rankloss}
\end{equation}
where recall $\delta_{ab}(t) = S(d_t, V_a) - S(d_t, V_b)$. In Eqn.~\ref{eq:rankloss}, $\gamma$ is a regularization factor, and $\varepsilon$ is used as a relaxation factor to set an upper bound to the loss to avoid it to return extremely large values. The loss $L_{f}(d_t)$ can be differentiated w.r.t. the parameters of the network $f$, and therefore the network can be learned through typical backpropagation methods. The training process is also illustrated in Fig.~\ref{fig:diagram}.
We want to recall that by using the Rank Loss function of Eqn.~\ref{eq:rankloss}, we are not backpropagating w.r.t. a defined ``ground-truth" $d^*_t$. In other words, our method is trained without the need of explicitly generating a target DR for each training image. This way, the network also contributes to define the form of the DR.





\subsection{Face analysis using Multi-level DR}
\label{subsec:apply}
\noindent We now describe how the DR shown above can be applied to face-related tasks. In particular, we observe that we can generate a multi-level set of DRs, each capturing a different temporal scale by using a different window length. We will validate that this combination allows to reach state of the art results in the tasks of AU intensity estimation and dimensional affect estimation. We first note that the generated $d_t$ are 3-channel tensors, no matter the choice of $T$. Thus, we can train a different model for different values of $T$, and combine the outputs before applying them to further related tasks. Herein, we will explore the use of a \textbf{Single Dynamic Representation} (\textbf{SDR}), using just the generated DR, and the use of a \textbf{Multi-level Dynamic Representation (\textbf{MDR})}, which combines the output of networks trained using different time lengths $T$. 

While there is no limit as to how many different levels can be used for further tasks, we want to keep the input network as simple as possible. To this end, in this paper we explore two different configurations, one for the AU intensity estimation and one for the dimensional affect estimation. We leave for future work to explore possible configurations. We rely on expert knowledge for each of the tasks: we choose a two-level representation for the AU intensity estimation task, and a three-level representation for the dimensional affect estimation task. In both, the first level corresponds to $T=0$, i.e. a window length of one frame. In practice, this level does not require the training of a DR, as it basically consists of the input image. Indeed, we experimentally validate that the use of the input image along with the DR helps capturing the rich appearance details in the input image along with the temporal dynamics given by the DR. For the dimensional affect estimation task, we choose $T=3$ for the second level. Finally, for both the second level of the AU intensity task and for the third level of the dimensional affect estimation task, we choose $T=5$. Note that the total window size is defined as $N = 2T+1$, i.e. $N = 7$ frames for $T=3$, and $N=11$ frames for $T=7$. In all cases, we use a stride of $S=2$ frames, i.e. we use every other frame to train the corresponding DR network. The chosen window then spans a set of $2TS + 1$ frames. This MDR yields a 6-channel tensor for the AU intensity task, and a 9-channel tensor for the dimensional affect estimation task. Fig.~\ref{fig:first} shows a description of the MDR for this three level approach.

\section{Experiments}
\label{sec:exp}
\noindent To validate the proposed approach, we first evaluate the ranking capability of the generated DRs. Then, we demonstrate their value for the tasks of AU intensity and dimensional affect estimation. We will show that our MDR produces state of the art results in both tasks. It is important to note that, at test time, a single face image is used as the input  to generate the corresponding MDR, and this representation is used to predict the values of the corresponding task. In addition, we want to remark that the MDR network is trained using the RECOLA dataset (see below), whereas the AU intensity network and the dimensional affect estimation network are trained using the BP4D and SEMAINE datasets, respectively. In other words, the DR networks are trained using a different dataset to those used to the corresponding tasks. 

\subsection{Datasets}
\label{ssec:data}
\noindent Experiments were conducted on three face datasets: RECOLA~\cite{ringeval2013introducing}, SEMAINE~\cite{mckeown2012semaine} and BP4D~\cite{zhang2014bp4d}. \textit{The RECOLA dataset is solely used to train the DR networks, whereas SEMAINE and BP4D are used to test both the capabilities of the DR to rank the corresponding frames, and to train and test the corresponding face-related tasks}. For the RECOLA dataset, we use the $27$ videos corresponding to the AVEC 2016 challenge~\cite{valstar2016avec}, each containing approximately $5$ minutes of people performing video conference. For the SEMAINE dataset, we use the subset predefined by the AVEC 2012 challenge~\cite{schuller2012avec}, which consists of $31$ training videos, $32$ validation videos, and $32$ test videos. All frames have been annotated with valence and arousal intensities, each lying in the range $[-1,1]$. For the BP4D dataset, we use the partitions predefined by the FERA 2015 challenge~\cite{valstar2015fera}. There are 75,586 frames for training, 71,260 frames for development and 75,726 frames in the test set. All the frames have been annotated for five AUs (AU6, AU10, AU12, AU14, and AU17), each lying in the range $[0,5]$. The sampling rate of RECOLA and BP4D was $25$ fps, whereas the sampling rate of SEMAINE was $\sim 50$ fps. For this reason, wherever we refer to the stride $S$ when setting up the span of frames to be considered, this will be automatically scaled to $2S$ for the SEMAINE dataset.

\subsection{Implementation details}
\label{ssec:setup}
\noindent \textbf{Configuration}  All the experiments are carried out using the PyTorch library~\cite{paszke2017pytorch} for deep learning. The network chosen for the task of generating the DRs is the UNet~\cite{ronneberger2015u}. Both the input and the output of the UNet are tensors of size $224\times224\times3$. The parameter $\theta$ in Eqn.~\ref{eq:rankloss} is set beforehand to ensure the chance level ranking accuracy to be less than $0.1\%$. For the task of AU recognition, we re-trained the network proposed by~\cite{sanchez2018joint}, as multi-task learning has been frequently adopted for AU recognition~\cite{linh2017deepcoder,nicolle2015facial}. For affect estimation, we followed the setting in~\cite{Kollias2019}, and fine-tuned the VGG-16 face network~\cite{parkhi2015deep}, whose last layer is modified to have an output size of $1$. 

\textbf{Pre-processing} We used the publicly available iCCR face tracker of~\cite{sanchez2018functional} to first detect a set of $66$ facial landmarks. Using these landmarks, images are cropped to meet the network size. Then, all pixels corresponding to the outer part of the convex hull defined by the landmarks are set to zero to remove all non facial appearance information.

\textbf{Training details} The UNet was trained using an Adam optimizer~\cite{kingma2015} with a learning rate of $10^{-3}$, and $\beta = (0.5, 0.9)$. During the ranking experiment, the UNet is trained and validated using the RECOLA dataset, and tested on all frames in the BP4D and SEMAINE datasets. For the face related tasks, we utilized the trained DR models to generate the DRs for each frame in the SEMAINE and BP4D datasets, and then trained the corresponding AU/dimensional affect models using the generated representations.

\subsection{Frame Ranking}
\label{ssec:ranking}
\noindent In this section, we report the ability of the generated DRs to rank the corresponding adjacent frames in SEMAINE and BP4D. In particular, we evaluate the capabilities of our model under different scenarios by choosing a set of different window lengths and strides to train the networks and generate the corresponding DRs. The number of frames used per training image is $N=2T+1$ ($T$ preceding frames, $T$ proceeding frames and the given frame). We sample $N$ frames using four different strides $S$. The image sequence range is then of $N \times S$ frames. We check the capabilities of our network for $T=\{3,5,7,9\}$ (i.e. $N = \{7,11,15,19\}$). In the most extreme case, i.e. when $T=9$ and $S=4$, the ranking is measured on a window size of $N=2T+1=19$ frames, evenly sampled from a sequence of $N\times S=76$ frames. At test time, frames are chosen following the same sampling procedure as that of the corresponding model. To compute the ranking accuracy, we compute the DR for each of the images available in the corresponding datasets. Then, we project the generated DR onto the frames lying within the corresponding window of $N$ frames, sampled with a stride $S$. We then measure the distances $\delta_{ab}(t)$ as defined in Eqn.~\ref{eq:score}, and measure the percentage of pairs that are correctly ranked, i.e. the percentage of pairs for which $\delta_{ab}(t) > 0$.

\begin{figure}
\centering
\includegraphics[width=8.6cm]{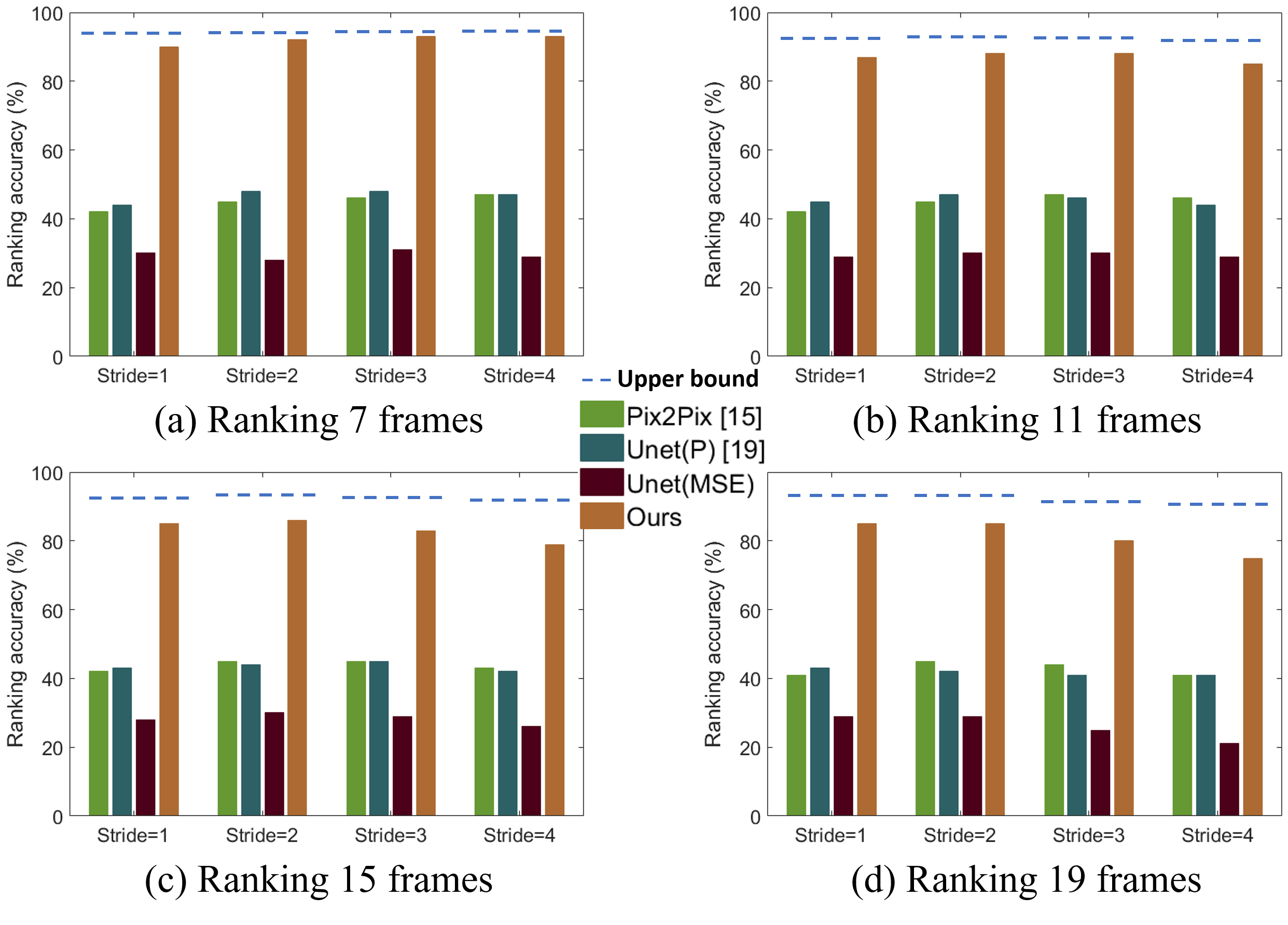}
\caption{Average ranking accuracy (\%) on two datasets. Four generative models are trained using RECOLA dataset and tested on SEMAINE and BP4D datasets. RankSVM classifiers were trained on SEMAINE and BP4D datasets, and each classifier only rank its training frames. The results obtained by RankSVM are treated as the upper bound.}
\label{fig:rank}
\end{figure}

In Fig.~\ref{fig:rank} we report the ranking accuracy of our method, measured as a percentage of correctly ranked frames w.r.t. total number of frames evaluated. We report the average accuracy measured across both SEMAINE and BP4D. The results shown with a dash line correspond to applying a RankSVM at test time, trained using the frames that were later ranked by it. The results given by the RankSVM are treated as an \textit{upper bound} for the ranking accuracy. 

We compare the accuracy of our method against methods trained using target representations obtained by the Dynamic Image algorithm, which generates an explicit target representation for each training image  using the RankSVM algorithm~\cite{bilen2017action}. We generated the corresponding dynamic images both forward (\textbf{DI}), and backward (\textbf{BDI}) in time, and applied them as the targets to train several generative networks. We trained a different network for the DI and the BDI images, respectively. In particular, Fig.~\ref{fig:rank} shows the results given by the following approaches:
\begin{itemize}
    \item UNet (MSE). Using the dynamic images, we train the model using as objective the Mean Squared Error. 
    \item UNet (P). In this method, we use the dynamic images as target representation, and the objective function used to train the model is the Perceptual loss proposed in~\cite{johnson2016perceptual}.
    \item Pix2Pix~\cite{Isola2017ImagetoImageTW} refers to using a conditional GAN, again using the dynamic images as the corresponding targets.
\end{itemize}

\begin{figure}
\centering
\includegraphics[width=8.6cm]{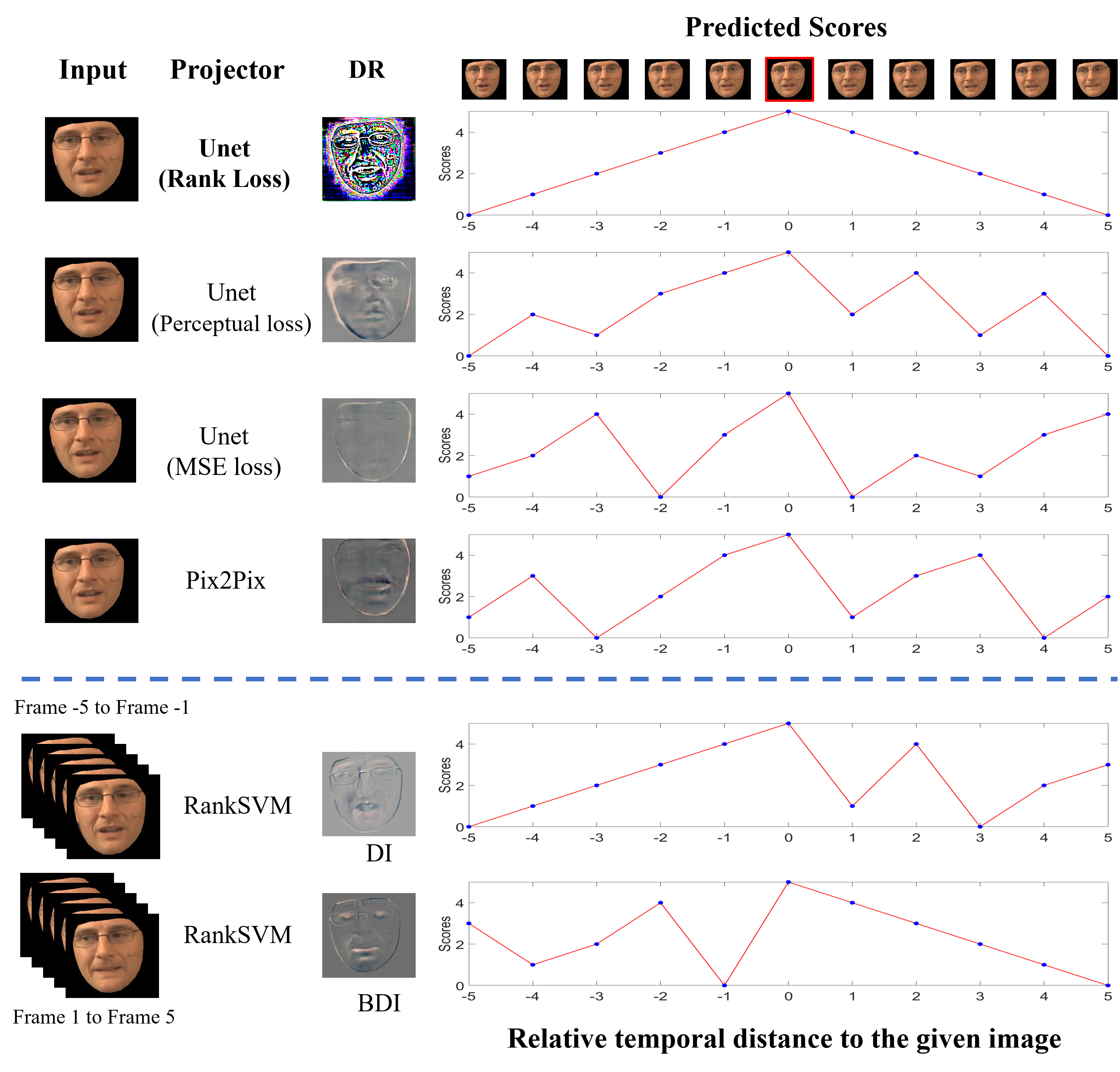}
\caption{Examples of ranking frames using DRs generated by different methods. The networks of Pix2Pix, Unet(P) and Unet(MSE) were trained using Seq DI as the target.}
\label{fig:result}
\end{figure}
The results shown in Fig.~\ref{fig:rank} show how our method achieves similar results to those given by the RankSVM, which uses at test time the adjacent frames to compute the kernel. Remarkably, our method yields around $80\%$ accuracy even for the longest cases (i.e. when ranking $19$ frames with different strides). In addition, we can see that, when pairing the input images with a DR to serve as a basis to learn our network, the ranking accuracy at test time degrades substantially. This illustrates the contribution of the Rank Loss at the task of defining the form of the DR, allowing for a better generalization. An example is shown in Fig.~\ref{fig:result}. It can be seen that our proposed DR is capable of accurately ranking both preceding and proceeding frames, something not possible with methods trained with explicit target representations. In addition, the RankSVM would only be able to rank preceding or proceeding frames, and it would need to be given the temporal information at test time. This clearly illustrates the capacities of our proposed approach at the task of sorting the surrounding frames of a given image relative to their position to it.  

\subsection{Face-related tasks}
\noindent In this section, we show the efficacy of the proposed approach for the tasks of AU intensity estimation and dimensional affect estimation, for which the temporal modeling would generally enhance the performance. As we demonstrate, our still image-based method yields similar results to those given by the models using image sequence, which make use of the temporal information. However, contrary to the temporal modeling of adjacent frames, our method only requires a single still image.

\paragraph{AU intensity estimation}
We first evaluate the contribution of our proposed approach at the task of AU intensity estimation. To do so, we re-train the network proposed by \cite{sanchez2018joint} using the training partition of BP4D. Then, we evaluate the performance of the proposed SDR and MDR approaches against using a single input image, as in \cite{sanchez2018joint}. We evaluate each method employing the same measures as those used by the FERA 2015~\cite{valstar2015fera} to rank participants: the Intra-Class Correlation (ICC(3,1),~\cite{shrout1979intraclass}), and the Mean Squared Error (MSE). We also compare the performance of our method against using the alternatives shown in Sec.~\ref{ssec:ranking}, i.e. those using an explicit target DR for training. Finally, we compare our approach against most recent works reporting state of the art results on the BP4D dataset. The results are shown in Table~\ref{BP4D}. 

From the results shown in Table~\ref{BP4D}, it can be seen that our MDR approach is capable of achieving state of the art results even when using as input a single image. In addition, we observe that the SDR achieved similar results to using a single image, i.e. the SDR approach gives similar results to those of ~\cite{sanchez2018joint}. We conjecture that while the SDR can encode the temporal pattern around a given frame, the original input image is rich in appearance details that remain important to infer the AU intensities. Thus, the best results are attained when combining the input image with the DRs (i.e. using the MDR approach). In other words, the estimated dynamics helped the still image-based AU intensity estimation to yield better results. 


\setlength{\tabcolsep}{4pt}
\begin{table}[t!]
	\begin{center}
      \scalebox{0.88}[0.88]{
		\begin{tabular}{|l| c| c c c c cl|}
			\toprule
 &AU & 6 & 10 & 12 & 14 &17 & Avg.   \\
\hline \hline
\multirow{10}*{ICC} 

&CCNN-IT~\cite{walecki2017deep} & 0.75 & 0.69 & 0.86 &0.40 &0.45&0.63 \\

&2DC~\cite{linh2017deepcoder} &0.76 & 0.71 &0.85 & 0.45& \textbf{0.53} & 0.66 \\

&VGP-AE~\cite{eleftheriadis2016variational} &0.75 & 0.66 & \textbf{0.88} &0.47& 0.49 & 0.65 \\

&HG-HMR \cite{sanchez2018joint} &\textbf{0.79} &0.80 &0.86 &0.54 &0.43 & 0.68 \\


&Pix2Pix$^{*}$ ~\cite{Isola2017ImagetoImageTW} & 0.59 & 0.62 & 0.68 &0.29 &0.31 & 0.50 \\

&Unet(P)$^{*}$ ~\cite{johnson2016perceptual} &0.55 & 0.65 & 0.65 &0.30& 0.26 & 0.48 \\

&Unet(MSE)$^{*}$  &0.56 & 0.63 & 0.66 &0.29& 0.26 & 0.48  \\

&\textbf{SDR}+HG-HMR &0.78 & 0.80 & 0.85 &0.47& 0.45 & 0.67 \\

&\textbf{MDR}+HG-HMR & 0.77 & \textbf{0.83} & 0.87  &\textbf{0.62} & 0.49 &\textbf{0.72} \\

\hline
\multirow{10}*{MSE} 

&CCNN-IT~\cite{walecki2017deep} & 1.23 & 1.69 & 0.98 &2.72 &1.17&1.57 \\

&2DC~\cite{linh2017deepcoder} &0.75 & 1.02 &0.66 & 1.44& 0.88 & 0.95 \\

&VGP-AE~\cite{eleftheriadis2016variational} &0.82 & 1.28 & 0.70 &1.43& 0.77 & 1.00 \\

&HG-HMR$^{*}$  \cite{sanchez2018joint} &\textbf{0.77} &0.92 &0.65 &1.57 &0.77&0.94 \\

&Pix2Pix$^{*}$ ~\cite{Isola2017ImagetoImageTW} &1.22 &1.31 &0.85 &1.90 &0.92 &1.24 \\

&Unet(P)$^{*}$ ~\cite{johnson2016perceptual} &1.53 &1.08 &1.07 &1.62 &0.95&1.25 \\

&Unet(MSE)$^{*}$  &1.09 &1.55 &1.18 &2.12 &1.15 &1.42\\

&\textbf{SDR}+HG-HMR &0.88 & 0.84 & 0.75 &1.90& 0.60 & 0.99 \\

&\textbf{MDR}+HG-HMR &0.99 & \textbf{0.79} & \textbf{0.64}& \textbf{1.34}& \textbf{0.48} & \textbf{0.85} \\
			\bottomrule
		\end{tabular}
        }
	\end{center}
	\caption{AU intensities estimation results on BP4D dataset. $*$ denotes results obtained by our own implementation} 
\label{BP4D}
\end{table}
\setlength{\tabcolsep}{1.4pt}

\paragraph{Dimensional affect estimation}
We measure the performance of our proposed approach at the task of dimensional affect estimation, i.e. at predicting the values of valence and arousal. To do so, we use the standard measures reported on the SEMAINE dataset, i.e. the Pearson Correlation Coefficient (PCC), and the Mean Squared Error (MSE). As introduced in Sec.~\ref{ssec:setup}, we fine-tune the VGG-16 network~\cite{parkhi2015deep} for each of the alternatives aforementioned. More specifically, we use a VGG-16 network for the inputs generated by our SDR and MDR, as well as for the methods trained using fixed representations, UNet(MSE), UNet(P), and Pix2Pix. In addition, we compare our approach against using a single image as input to the VGG-16 (SI+VGG). The results are shown in Table~\ref{SEMAINE_other method}. Again, we compare against state of the art methods on valence and arousal estimation using visual information, including those that make use of temporal information to improve their performance (\cite{savran2012combining,kaltwang2016doubly}). As it can be seen, our proposed MDR approach attains state of the art results using as input only still images. 



\setlength{\tabcolsep}{4pt}
\begin{table}[t!]
	\begin{center}
      \scalebox{1}[1]{
		\begin{tabular}{|l| c c c c c |l|}
			\toprule
			&
			\multicolumn{2}{c}{Arousal}&&
			\multicolumn{2}{c|}{Valence}\\
			\hline\hline
Method & PCC & MSE &  & PCC &MSE  \\
			\midrule
Savran et al.$^{\dagger}$ \cite{savran2012combining} &0.251 & N.A. & &0.210& N.A. \\
Kaltwang et al.$^{\dagger}$ \cite{kaltwang2016doubly} & 0.310 & \textbf{0.042} & &0.310 & \textbf{0.058} \\
\hline
Zhang et al. \cite{zhang2014representation} & 0.070  &N.A. & &0.241&N.A. \\
SI+VGG $^{*}$ ~\cite{parkhi2015deep}& 0.246 & 0.056 & & 0.258 &0.084 \\
Pix2Pix$^{*}$+VGG~\cite{Isola2017ImagetoImageTW} & 0.091 & 0.166 & &0.088 &0.195 \\
Unet(P)$^{*}$+VGG~\cite{johnson2016perceptual}&0.192 & 0.125 & &0.134& 0.172 \\
Unet(MSE)$^{*}$+VGG&0.063 & 0.181 & &0.082& 0.189 \\
\hline
\textbf{SDR}+VGG & 0.306 & 0.078  & & 0.299 & 0.082 \\
\textbf{MDR}+VGG & \textbf{0.335} & 0.058  & & \textbf{0.316} & 0.072 \\
			\bottomrule
		\end{tabular}
        }
	\end{center}
	\caption{Affect estimation results on the SEMAINE dataset. SI denotes the still face image; $*$ denotes our own implementation; $\dagger$ denotes methods that rely on the use of temporal information at test time}
\label{SEMAINE_other method}
\end{table}
\setlength{\tabcolsep}{1.4pt}

\section{Conclusion}
\noindent In this paper, we have proposed a novel approach to model temporal dynamics from static face images. This approach allows us to train a network to infer a DR for a previously unseen test image, which effectively summarize dynamics surround it. We illustrated that the generated DRs can be used indistinctly for the tasks of Action Unit intensity and dimensional affect estimation, attaining state of the art results. We empirically validated the capacity of the DRs to rank unseen frames in test time, as well as their contribution to the face-related tasks. In addition, we validated that a network trained with a rank loss function generalizes better to unseen images than a model trained using pre-defined representations, i.e. we demonstrated the ability of our network to be properly learned without the need of target representations. Experimental results have shown that our method is powerful not only for the task of ranking adjacent frames in different facial actions but also for the tasks of estimating the AU and affect intensities from the face. 

{\small
\bibliographystyle{ieee}
\bibliography{egbib}
}

\end{document}